\pdfoutput=1

\documentclass[11pt]{article}

\usepackage[final]{acl}

\usepackage{times}
\usepackage{latexsym}

\usepackage[T1]{fontenc}

\usepackage[utf8]{inputenc}

\usepackage{microtype}

\usepackage{inconsolata}

\usepackage{graphicx}
\usepackage[skip=0pt]{subcaption}
\usepackage{multirow}
\usepackage{xcolor}
\usepackage{newverbs}
\newverbcommand{\cverb}{\color{red}}{}
\newverbcommand{\bverb}
  {\begin{lrbox}{\verbbox}}
  {\end{lrbox}\colorbox{gray!30}{\box\verbbox}}

\makeatletter
\def\thickhline{%
  \noalign{\ifnum0=`}\fi\hrule \@height \thickarrayrulewidth \futurelet
   \reserved@a\@xthickhline}
\def\@xthickhline{\ifx\reserved@a\thickhline
               \vskip\doublerulesep
               \vskip-\thickarrayrulewidth
             \fi
      \ifnum0=`{\fi}}
\makeatother

\newlength{\thickarrayrulewidth}
\usepackage{listings,multicol}  
\usepackage{filecontents}       
\usepackage{lipsum}             

\usepackage{capt-of}
\usepackage{rotating}
\usepackage{float}
\usepackage{booktabs,amsfonts,dcolumn}
\newcolumntype{d}[1]{D..{#1}}

\usepackage{algorithm}
\usepackage{algpseudocode}
\usepackage{amsmath,amssymb,amsfonts}
\usepackage{amsthm}
\usepackage{mathtools}

\algnewcommand{\LeftComment}[1]{\Statex \(\triangleright\) #1}
\DeclareMathOperator*{\argmax}{\arg\max}

\newtheorem*{assumption*}{\assumptionnumber}
\providecommand{\assumptionnumber}{}


\makeatletter
\renewcommand\paragraph{\@startsection{paragraph}{4}{\z@}%
	{1.5ex plus .2ex minus .3ex}%
	{-0em}%
	{\normalsize\bf}}
\makeatother

\usepackage{enumitem}
\setlist{nosep,  
  align=left,
  left=0pt,}

\usepackage{authblk}
\title{HGOT: Hierarchical Graph of Thoughts for Retrieval-Augmented In-Context Learning in Factuality Evaluation}

\author[1,3]{Yihao Fang}
\author[1]{Stephen W. Thomas}
\author[2,3]{Xiaodan Zhu}
\affil[1]{Smith School of Business, Queen’s University}
\affil[2]{Department of Electrical and Computer Engineering, Queen's University}
\affil[3]{Ingenuity Labs Research Institute, Queen's University \authorcr
yihao.fang@gmail.com,
 \{stephen.thomas, xiaodan.zhu\}@queensu.ca}

\usepackage{fancyhdr}

\fancypagestyle{Proceedings}{\fancyhf{} \fancyfoot[C]{Proceedings of the Fourth Workshop on Trustworthy Natural Language Processing (TrustNLP 2024), Mexico City, Mexico. Association for Computational Linguistics.} }

\begin{document}
\thispagestyle{Proceedings}
\maketitle
\begin{abstract}
With the widespread adoption of large language models (LLMs) in numerous applications, the challenge of factuality and the propensity for hallucinations has emerged as a significant concern. To address this issue, particularly in retrieval-augmented in-context learning, we introduce the hierarchical graph of thoughts (HGOT), a structured, multi-layered graph approach designed to enhance the retrieval of pertinent passages during in-context learning. The framework utilizes the emergent planning capabilities of LLMs, employing the divide-and-conquer strategy to break down complex queries into manageable sub-queries. It refines self-consistency majority voting for answer selection, which incorporates the recently proposed citation recall and precision metrics to assess the quality of thoughts, linking an answer's credibility intrinsically to the thought's quality. This methodology introduces a weighted system in majority voting, prioritizing answers based on the citation quality of their thoughts. Additionally, we propose a scoring mechanism for evaluating retrieved passages, considering factors such as citation frequency and quality, self-consistency confidence, and the retrieval module's ranking. Experiments indicate that HGOT excels as a versatile approach, outperforming competing models in FEVER by up to $7\%$ and matching leading models such as Retrieve-then-Read in Open-SQuAD, and DSP in HotPotQA, demonstrating its efficacy in enhancing LLMs' factuality.
\end{abstract}

\section{Introduction}

The advancement of large language models (LLMs) \citep{devlin-etal-2019-bert, raffel2020exploring, radford2018improving, radford2019language, brown2020language} has revolutionized the field of NLP and artificial intelligence by offering unprecedented capabilities in natural language understanding and generation, leading to their widespread adoption in many applications. However, a critical challenge of these models is the tendency to ``hallucinate'' \citep{maynez-etal-2020-faithfulness, raunak-etal-2021-curious, bouyamourn-2023-llms}---generating content that is factually incorrect or not grounded in reality. This issue raises significant concerns about the reliability and trustworthiness of LLMs, particularly in high-stakes applications. While numerous efforts have been made to address various aspects of this problem, a specific area that demands attention is retrieval-augmented in-context learning \citep{lazaridou2022internet, izacard2022few, press2022measuring, khattab2022demonstrate}, a process where LLMs leverage external information to enhance their responses.

In response to the challenge of hallucinations, we introduce the hierarchical graph of thoughts (HGOT) framework, drawing inspiration from neuropsychological studies on the ``hierarchy of goals'' and working memory \citep{cowan2010multiple, jonides2008mind, cowan2005working}. Our approach redefines how LLMs interact with and utilize external information sources. By constructing a structured, multi-layered graph \citep{ying2018hierarchical, chen2021hierarchical}, HGOT allows for a more organized and efficient way of sourcing and incorporating relevant information, thereby reducing the incidence of hallucinations in LLMs. Despite these advances, the challenges that we need to overcome involve dynamically constructing a hierarchical graph, as well as evaluating and ranking the qualities of thoughts and retrieved passages in this complex structure. 

The HGOT framework places a strong emphasis on the dynamic creation of a hierarchical graph structure by exploring the applicability of the emergent planning capabilities of LLMs \citep{wang2023plan, valmeekam2023planning} in breaking down complex queries (higher in the hierarchy) into simpler sub-queries (lower in the hierarchy). This method employs a divide-and-conquer strategy, which simplifies the problem-solving process and improves the accuracy and relevance of the information retrieved by the LLM. 

Another key feature of the HGOT framework is the improvement of the self-consistency majority voting mechanism \citep{wang2023selfconsistency} used in LLMs, which enhances the quality assessment of thoughts or rationales. This improvement assesses the quality of thoughts or rationales generated by the LLMs. The method utilizes metrics such as citation recall and precision \citep{gao2023enabling} to evaluate the quality of the information used by the LLMs in forming their responses. The underlying premise is that the quality of an LLM's response is directly related to the quality of its underlying thought. Therefore, in the majority voting process, responses are given weights based on the citation quality of their thoughts. 

Furthermore, the HGOT framework proposes a scoring mechanism to evaluate the quality of retrieved passages. This mechanism takes into account various factors, including the frequency of passage citation, the citation quality \citep{gao2023enabling} of the thought, self-consistency confidence score \citep{xiong2023can, wang2023selfconsistency}, and the retrieval module ranking. By considering these diverse factors, the mechanism ensures that the information utilized in the LLM's response generation is both relevant and of high quality.

To validate the effectiveness of the proposed method, we selected FEVER \citep{thorne-etal-2018-fever}, Open-SQuAD \citep{rajpurkar-etal-2016-squad, karpukhin-etal-2020-dense}, and HotPotQA \citep{yang-etal-2018-hotpotqa} to evaluate the models' proficiency in fact retrieval and reasoning. We divided these datasets into three groups: ``Long'', ``Medium'', and ``Short'', according to the question length, emphasizing sampling from the tails of the distribution, a detail that is frequently overlooked in studies. Experiments show that HGOT outperforms existing retrieval-augmented in-context learning methods in FEVER by up to $7\%$ and matching leading models such as Retrieve-then-Read \citep{lazaridou2022internet, izacard2022few} in Open-SQuAD, and Demonstrate-Search-Predict (DSP) \citep{khattab2022demonstrate} in HotPotQA, underscoring its robustness and efficacy in enhancing LLMs' factuality.

In brief, we make the following contributions: 
\begin{itemize}
\item
We introduce HGOT and investigate LLM's (emergent) planning ability in breaking down complex queries for graph construction.

\item
\textbf{Thought Quality:} HGOT selects the best answer by voting which involves assessing thought quality with citation recall and precision metrics.

\item
\textbf{Retrieval Quality:} We propose a scoring mechanism for evaluating retrieved passages based on citation frequency and quality, self-consistency confidence, and retrieval module ranking.

\item 
We conduct extensive experiments on FEVER, Open-SQuAD, and HotPotQA, emphasizing sampling from the extremes of the distribution. The results demonstrate HGOT's efficacy in enhancing LLMs' factuality.
\end{itemize}

\section{Related Work}

The ``Retrieve-then-Read'' pipeline \citep{lazaridou2022internet, izacard2022few} sends queries to a retrieval model (RM) to gather passages for a prompt that a language model (LM) uses for response generation. ``Self-ask'' \citep{press2022measuring} and ``Iterative
Retriever, Reader, and Reranker'' (IRRR) \citep{qi2020answering} improve upon this approach through multi-hop retrieval, enabling the LM to ask follow-up questions that the RM answers. These answers, combined with the original prompt, enhance the LM's ability to respond to the initial question.



``ReAct'' \citep{yao2023react} uses LLMs to generate reasoning traces and task-specific actions in an interleaved manner. While reasoning traces help the model induce, actions allow it to interface with external sources. Baleen \citep{khattab2021baleen} summarizes multiple passages of information in each hop to be used in subsequent iterations. The ``Demonstrate-Search-Predict'' (\texttt{DSP}) approach \citep{khattab2022demonstrate} enhances the multi-hop methodologies by automatically annotating ``chain-of-thought'' \citep{wei2022chain} demonstrations. The potential weakness of those multi-hop pipelines lies in the generality and adaptability of their search operations. Especially, those pipelines face challenges when tasked with addressing inquiries that necessitate intricate planning for the retrieval of pertinent information. 

\begin{figure*}
\centering
  \includegraphics[width=0.86\linewidth]{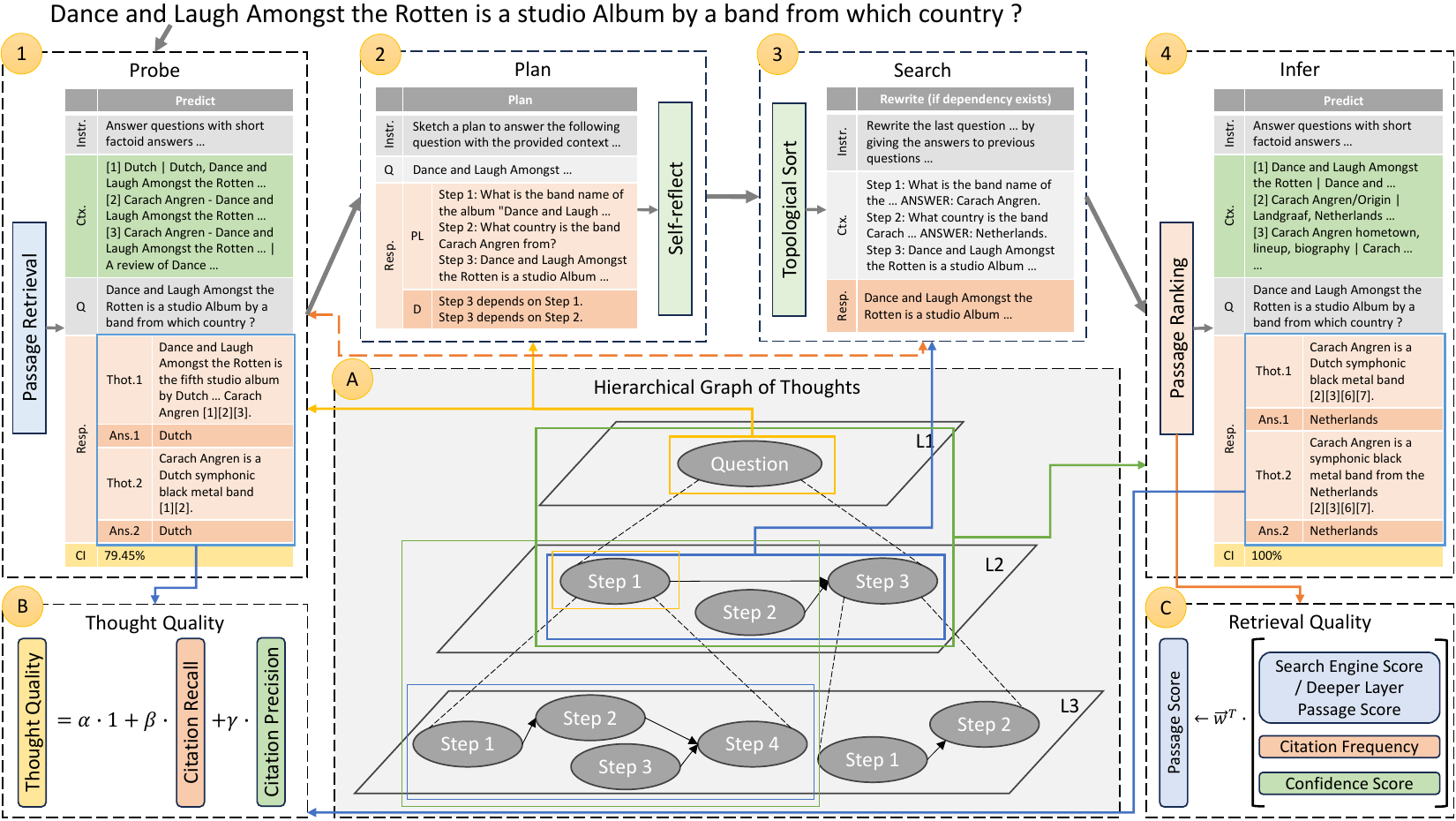}
  \caption{An illustrative example of HGOT in answering a factual question. (The abbreviations employed are as follows: Instr.: Instructions, Q: Question, Ctx.: Context or References, Resp.: ChatGPT's Response, PL: Plan, D: Dependencies, CI: Confidence, Ans.: Answer, Thot.: Thought)}
  \label{fig:methodology}
\end{figure*}

Plan-and-Solve (PS) Prompting \citep{wang2023plan} involves breaking down complex tasks into manageable subtasks and executing them according to a formulated plan, with PS+ prompting enhancing reasoning quality through detailed instructions. However, PS hasn't yet utilized LLMs' planning capabilities with retrieval-augmented in-context learning. Other methods such as the ``tree of thoughts'' \citep{yao2023tree}, ``graph of thoughts'' \citep{besta2023graph}, and RECURRENTGPT \citep{zhou2023recurrentgpt} explore reasoning via tree, graph, or recurrent structures to improve problem-solving, but they face challenges in sourcing relevant information, suffering from drawbacks concerning the factual reliability of large language models. 




\section{Methodology}

The HGOT framework involves creating a multi-layered graph that allows for a more organized and efficient sourcing and incorporation of relevant information. This structure aims to reduce the occurrence of hallucinations in LLMs. However, the initial challenges that we need to overcome involve dynamically constructing hierarchical graphs, along with assessing and ranking the qualities of thoughts and retrieved passages within this complex structure.

In terms of hierarchical graph construction, the HGOT framework utilizes the emergent planning ability of LLMs to break down complex queries into smaller, more manageable sub-queries (or steps), following a divide-and-conquer strategy. 

To select the best answer for a query, the framework employs a method of improving self-consistency majority voting \citep{wang2023selfconsistency}. This involves assessing the quality of thoughts using citation recall and precision metrics and weighing answers based on the citation quality of their thoughts (Figure~\ref{fig:methodology}: \raisebox{.5pt}{\textcircled{\raisebox{-.9pt} {B}}}).

Additionally, a scoring mechanism is proposed for evaluating the quality of retrieved passages. This mechanism takes into account various factors such as the frequency of passage citation, the quality of citations in the thoughts, a self-consistency confidence score adjusted for citation quality, and the retrieval module's ranking (Figure~\ref{fig:methodology}: \raisebox{.5pt}{\textcircled{\raisebox{-.9pt} {C}}}).

\subsection{Hierarchical Graph Construction, Search, and Inference}

\paragraph{Graph Construction: } When utilizing the emergent planning ability to break down a complex question into smaller, more manageable sub-queries or steps, it's crucial to recognize that these sub-queries or steps are not standalone. Instead, they often exhibit interconnections that contribute to forming a complete answer. 
These steps and their connections create a dependency graph within a deeper level of the hierarchical graph, which guides the exploration of the complex question. (In this framework, the dependency graph is designed as a directed acyclic graph to avoid circular dependencies.)
 Further, each sub-query can be extended into a more detailed dependency graph at even deeper levels of the hierarchy. For example, as illustrated in Figure~\ref{fig:methodology}: \raisebox{.5pt}{\textcircled{\raisebox{-.9pt} {A}}}, a query at the initial layer (Layer 1 or L1) can be extended into a dependency graph at a subsequent layer (Layer 2 or L2). Within L2, the first step could unfold into a four-step dependency graph in the next layer (Layer 3 or L3), while the third step in L2 might lead to a two-step dependency graph at the same third layer (L3).

Establishing a precise dependency graph is essential before progressing to the subsequent stage, as any error or ambiguity at this stage could significantly derail the solution path. To accurately infer this graph, there are several strategies that we can adopt. Initially, we employed the ``Probe'' procedure to gather references (referenced in Figure~\ref{fig:methodology}: \raisebox{.5pt}{\textcircled{\raisebox{-.9pt} {1}}} and Appendix~\ref{sec:predict_prompt_and_response}). This involves collecting passages from the retrieval model and then scoring these passages by prompting LLM to probe for an answer. The specifics of how passages are scored will be discussed in Section~\ref{sec:retrieval_quality}. 

Subsequently, we designed the prompt template for the ``Plan'' procedure (Figure~\ref{fig:methodology}: \raisebox{.5pt}{\textcircled{\raisebox{-.9pt} {2}}} and Appendix~\ref{sec:plan_prompt_and_response}). This template incorporates instructions, demonstrations (see Appendix~\ref{sec:automated_annotated_demonstrations}), and the collected passages. The aim is to stimulate the LLM and guide it towards a holistic understanding of the question and its interconnected components.

Once the ``Plan'' procedure is complete, we introduce the self-reflection technique (Appendix~\ref{sec:self_reflect_prompt_and_response}), inspired by the work of  \citet{shinn2023reflexion}.
This involves prompting the LLM again to double-check if the output dependencies are accurate and align with the question in each step. The method encourages the LLM to focus internally on the dependencies without external influence, by providing only related steps or sub-queries. Finally, we formalize these dependencies into a structure that is more compatible with programming language formats (Appendix~\ref{sec:formalize_prompt_and_response}).

\paragraph{Search: } A crucial aspect of this stage involves using topological sorting and rewriting, as shown in Figure~\ref{fig:methodology}: \raisebox{.5pt}{\textcircled{\raisebox{-.9pt} {3}}}. Topological sorting within a dependency graph (i.e., a directed acyclic graph) ensures that steps influencing subsequent steps are processed in a sequential order. When evaluating a step or a sub-query, a ``Probe'' procedure is employed (refer to Figure~\ref{fig:methodology}: \raisebox{.5pt}{\textcircled{\raisebox{-.9pt} {1}}}), which gathers passages from the retrieval model and instructs the LLM to search for an answer by using the sub-query. In the context of the dependency graph, when Step~2 is contingent on Step~1, the question in Step~2 is rewritten (see Appendix~\ref{sec:rewrite_prompt_and_response}) to include the sub-query from Step~1 along with the answer obtained from the ``Probe'' procedure. This process ensures that the interconnections are well-articulated and traceable within the graph.

The ``Probe'' procedure for each sub-query does more than seek answers; it also gathers and scores relevant passages. Additionally, the ``Plan'' procedure is applied to each sub-query to create a dependency graph at a deeper level. Following this, the ``Search'' procedure (Figure~\ref{fig:methodology}: \raisebox{.5pt}{\textcircled{\raisebox{-.9pt} {3}}}) investigates the dependency graph topologically, and the ``Infer'' procedure (Figure~\ref{fig:methodology}: \raisebox{.5pt}{\textcircled{\raisebox{-.9pt} {4}}}) is then utilized to calculate the final scores for all the passages collected in the earlier stages, to predict the answer, and to determine the confidence score. In each step or sub-query assessed during the ``Search'' procedure, the ``Probe'',  ``Plan'', ``Search'', and ``Infer'' procedures are recursively executed until a specified depth of the graph is achieved, or the ``Plan'' procedure opts to stop further progression. Specifically, the termination condition is activated if the ``Plan'' procedure results in only a single step that closely resembles the sub-query being planned. The similarity between them is assessed using the cosine similarity of their BERT-based sentence embeddings \citep{reimers-gurevych-2019-sentence}.

\begin{algorithm}
  \small
  \caption{HGOT Traversal}\label{alg:approach}
  \begin{algorithmic}[1]
    \LeftComment{Let $q$ be a question}
    \LeftComment{Let $a$ be an answer. e.g., $a_q$ is the answer to $q$}
    \LeftComment{Let $\mathbf{G}$ be a dependency graph (i.e., a directed acyclic graph)}
    \LeftComment{Let $\mathbf{CTX}$ be the context (incl. passages and scores)}
    \LeftComment{Let $\mathbf{CI}$ be a confidence score}
    \LeftComment{Let $d$ be the level of depth in the hierarchical}
    \State
    \Procedure{TRAVERSE}{$q, d$}  
      \State $a_q, \mathbf{CI}_q, \mathbf{CTX}_q \leftarrow \text{PROBE}(q)$
      \State $\mathbf{G} \leftarrow \text{PLAN}(q, \mathbf{CTX}_q)$
      \If {$\text{STOP}(q, \mathbf{G}, d)$}
        \State \textbf{return} $a_q, \mathbf{CI}_q, \mathbf{CTX}_q$
      \Else
      \State $\mathbf{CTX}_{\mathbf{G}} \leftarrow \text{SEARCH}(\mathbf{G}, d+1)$
      \State $a_q, \mathbf{CI}_q, \mathbf{CTX} \leftarrow \text{INFER}(q, \mathbf{CTX}_q, \mathbf{CTX}_{\mathbf{G}})$
      \State \textbf{return} $a_q, \mathbf{CI}_q, \mathbf{CTX}$
      \EndIf
    \EndProcedure
    \State
    \Procedure{SEARCH}{$\mathbf{G}, d$} 
      \State $q_1, ..., q_r \leftarrow \text{TOPOLOGICAL\_SORT}(\mathbf{G})$
      \For {$i$ \textbf{in} $1...r$}
        \State $q_i \leftarrow \text{REWRITE}(q_i, \text{IN\_NEIGHBORS}(q_i, \mathbf{G}))$
        \State $a_{q_i}, \mathbf{CI}_{q_i}, \mathbf{CTX}_{q_i} \leftarrow \text{TRAVERSE}(q_i, d)$
      \EndFor
      \State \textbf{return} $\mathbf{CTX}_{q_1}, ..., \mathbf{CTX}_{q_r}$
    \EndProcedure
  \end{algorithmic}
  \label{alg:hgot}
\end{algorithm}

\paragraph{Inference: } Having the hierarchical graph of thoughts and their related passages collected from the retrieval model, the ``Infer'' procedure predicts the final answer to the query (Figure~\ref{fig:methodology}: \raisebox{.5pt}{\textcircled{\raisebox{-.9pt} {4}}}).
Specifically, this procedure ranks all passages retrieved during the examination of the query and its sub-queries, as will be explained in Section~\ref{sec:retrieval_quality}. It subsequently selects the top K passages with the highest rankings to use as the prompt for LLM. Along with demonstrations and instructions, the ``Infer'' procedure asks LLM to think step by step, predicts the final answer, and estimates the confidence score (Appendix~\ref{sec:predict_prompt_and_response} and Appendix~\ref{sec:automated_annotated_demonstrations}). The algorithm for recursive planning, searching, and inferring within HGOT is detailed in Algorithm~\ref{alg:hgot}.

\subsection{Thought Quality}
\label{sec:thought_quality}

When assessing the quality of thoughts, we establish tuples $(\tau_1, a_1), ..., (\tau_m, a_m)$ as pairs of LLM-generated thoughts (rationales) and answers, as shown in Figure~\ref{fig:methodology}: \raisebox{.5pt}{\textcircled{\raisebox{-.9pt} {1}}}, \raisebox{.5pt}{\textcircled{\raisebox{-.9pt} {4}}}, and \raisebox{.5pt}{\textcircled{\raisebox{-.9pt} {B}}}. The quality of a thought $\tau_i$ is determined by modifying the concepts of citation recall ($\text{REC}$) and citation precision ($\text{PREC}$) as introduced by \citet{gao2023enabling}, in the following manner:

\begin{equation}
    \rho_i \coloneqq \alpha \cdot 1 + \beta \cdot \text{REC}(\tau_i) + \gamma \cdot \text{PREC}(\tau_i)
\label{eq:thought_quality}
\end{equation}

Assuming there are $d$ distinct responses $\hat{a}_1, ..., \hat{a}_d$, with $d$ being less than or equal to $m$, we improve upon the self-consistency majority voting method \citep{wang2023selfconsistency} by factoring in the thought qualities, defining the selected answer as:

\begin{equation}
\hat{a}^* = \argmax_{\hat{a}_h \in \{\hat{a}_1, ..., \hat{a}_d\}} \sum^m_{i=1}  \rho_i \delta(a_i,\hat{a}_h)
\end{equation}
where $\delta$ is the Kronecker delta function, which equals 1 when the variables are the same and 0 otherwise.

Moreover, we develop the self-consistency confidence score \citep{xiong2023can} by taking into account the thought qualities. This is defined as:

\begin{equation}
\mathbf{CI} = \frac{\sum^m_{i=1}  \rho_i \delta(a_i,\hat{a}^*) }{\sum^m_{i=1} \rho_i}
\end{equation}

Note that when $\alpha$ equals 1 and both $\beta$ and $\gamma$ are zero, these equations are simplified to the prediction and calibration based on self-consistency \citep{wang2023selfconsistency,xiong2023can}.

\subsection{Retrieval Quality}
\label{sec:retrieval_quality}

Assessing the quality of retrieved passages considers multiple aspects. These include how often the passage is cited, the quality of these citations \citep{gao2023enabling}, a self-consistency confidence score \citep{xiong2023can}, and the ranking given by the retrieval module (Figure~\ref{fig:methodology}: \raisebox{.5pt}{\textcircled{\raisebox{-.9pt} {C}}}).

Assume $p$ is a particular passage retrieved, which serves as a part of the context in the ``Probe'' or ``Infer'' procedures. The pairs $(\tau_1, a_1), ..., (\tau_m, a_m)$ represent the generated thoughts (rationales) and answers produced when using ChatGPT with a temperature greater than zero. Statements or sentences $s_1, ..., s_{l_{\tau_i}}$ are  parts of $\tau_i$. The process of natural language inference (denoted as a function NLI) and a citation marker at the end of each statement (denoted as M) work together to determine if a statement $s_j$ cites passage $p$, resulting in a value of either true or false. This is formally expressed as:

\begin{equation}
    \hat{\delta}(p,s_j)=
    \begin{cases}
   1,& \text{if } \text{M}(p, s_j) \text{ or } \text{NLI}(p, s_j)\\
    0,              & \text{otherwise}
\end{cases}
\end{equation}

We further define the ``weighted citation frequency per thought'' for a given passage $p$, as the total number of citations in $\tau_i$, adjusted by the quality of the thought $\tau_i$. Formally, it is presented as:

\begin{equation}
\nu(p, \tau_i) = \rho_i \sum_{j=0}^{l_{\tau_i}} \hat{\delta}(p,s_j)
\end{equation}

The ``weighted citation frequency'' is the aggregate of these ``weighted citation frequencies per thought'' across all thoughts, and is denoted by:

\begin{equation}
\hat{\nu}(p) = \sum_{i=0}^m \nu(p, \tau_i)
\end{equation}

Next, we normalize this ``weighted citation frequency'' so that the highest value among all passages from a specific retrieval $P$, to which $p$ belongs, is equal to $1$. The ``normalized weighted citation frequency'' is thus:

\begin{equation}
\bar{\nu}(p)=\frac{\hat{\nu}(p)}{\max_{p \in P} \hat{\nu}(p)}
\end{equation}

Finally, during the ``Probe'' or ``Infer'' procedures, the quality score of the passage $p$ is updated repetitively, starting with the initial score $\sigma(p, 0)$ provided by the search engine in the ``Probe'' procedure. The formula is expressed as follows:

\begin{equation}
    \sigma(p, t+1) \leftarrow \vec{w}^T \cdot  \left[ {\begin{array}{cc}
    \sigma(p, t) \\
   \bar{\nu}(p)  \\
   \mathbf{CI}  \\
  \end{array} } \right]
  \label{eq:retrieval_quality}
\end{equation}
where $\vec{w}=(w_1, w_2, w_3)$ is a hyperparameter vector that can be tuned for different datasets, retrieval models and large language models.

\section{Data}

 We evaluate HGOT across three datasets: FEVER \citep{thorne-etal-2018-fever}, Open-SQuAD \cite{rajpurkar-etal-2016-squad,karpukhin-etal-2020-dense}, and HotPotQA \cite{yang-etal-2018-hotpotqa}. Considering the use of sentence length as a parameter for estimating complexity has been implemented in various NLP tasks \citep{platanios-etal-2019-competence,spitkovsky-etal-2010-baby}, to assess HGOT across different complexity levels, we stratify the three datasets based on sentence length, categorizing them into long, medium, and short.

\begin{figure}[h]
    \centering
    \includegraphics[width=\linewidth]{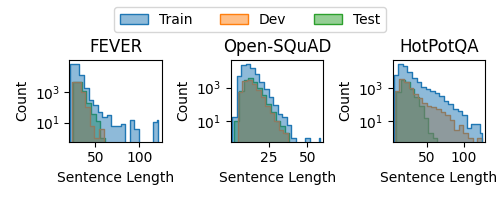}
    \caption{The sentence length, measured by the number of tokens in a question, from the FEVER, Open-SQuAD, and HotPotQA datasets}
    \label{fig:sentence_length}
\end{figure}

The sentence length, measured by the number of tokens in a question, from the FEVER, Open-SQuAD, and HotPotQA datasets is illustrated in Figure~\ref{fig:sentence_length}. The median number of tokens in FEVER is $27$, with a long tail of instances extending beyond the median (indicating possible complexity in reasoning, see Appendix~\ref{sec:dataset_examples_and_examination} for a more in-depth examination of the data). Open-SQuAD and HotPotQA likewise exhibited a similar distribution. The training, development, and test distributions align well with each other, enabling the stratification of these datasets by sentence length.

\renewcommand{\arraystretch}{1}
\begin{table}[h]
\small
\centering
\setlength\tabcolsep{1.5pt}
\begin{tabular}{l|rrr|rrr|rrr}
\thickhline
 \multirow{2}{0.1\linewidth}{\bf Sent. Len.}&\multicolumn{3}{c|}{\bf FEVER} &\multicolumn{3}{c|}{\bf Open-SQuAD}&\multicolumn{3}{c}{\bf HotPotQA}\\
\cline{2-10}
&\bf Train&\bf Dev&\bf Test&\bf Train&\bf Dev&\bf Test&\bf Train &\bf Dev&\bf Test\\
\hline
Long& 1619& 113& 113& 1174& 121& 118& 1504& 168& 137\\
Medium& 2182& 150& 150& 1181& 133& 159& 1628& 181& 148\\
Short& 2182& 150& 150& 1181& 133& 159& 1628& 181& 148\\
\thickhline
\end{tabular}

\caption{Count of examples across all three datasets and nine categories (Refer to Appendix~\ref{sec:dataset_summary_statistics} for summary statistics and Appendix~\ref{sec:dataset_examples_and_examination} for data examples)}
\label{tab:count_of_examples}
\end{table}
\renewcommand{\arraystretch}{1}

Questions from FEVER and Open-SQuAD that exceed the $98.5^{th}$ percentile in length are categorized as long, while for HotPotQA, this categorization applies to questions above the $98^{th}$ percentile. For questions of FEVER and Open-SQuAD that fall between the $1.5^{th}$ and $98.5^{th}$ percentiles, they are defined as medium length, and for HotPotQA, this range is from the $2^{nd}$ to the $98^{th}$ percentile. Within this group of medium-length questions, about $1.5\%$ of those from FEVER and Open-SQuAD are randomly chosen for evaluation, compared to $2\%$ of HotPotQA questions. Additionally, questions from FEVER and Open-SQuAD below the $1.5^{th}$ percentile are labelled as short, similar to those under the $2^{nd}$ percentile for HotPotQA questions. 
Lastly, Table~\ref{tab:count_of_examples} displays the total number of examples across all three datasets, spanning nine categories. 

\paragraph{Metrics: }  

For  Open-SQuAD and HotPotQA, we utilize the Exact Match (EM) and F1 scores \citep{rajpurkar-etal-2016-squad}. The EM score identifies the proportion of predictions that precisely align with the correct answers, while the F1 score assesses the average token overlap between the prediction and the correct answer. For FEVER, we only use the EM score, considering the answers in FEVER being limited to three tokens or fewer.

\section{Evaluation Setup}

\paragraph{Baselines: } Our benchmarking includes five approaches: ``Vanilla LM'' \citep{brown2020language}, ``Retrieve-then-Read'' \citep{lazaridou2022internet, izacard2022few}, ``Self-ask'' \citep{press2022measuring}, ``ReAct'' \citep{yao2023react}, and ``Demonstrate-Search-Predict'' (DSP) \citep{khattab2022demonstrate}. See Appendix~\ref{sec:baselines} for further details.

\paragraph{Implementation Details: } All approaches employed ChatGPT (gpt-3.5-turbo-1106) as the backbone LLM, with the exception of ReAct, which utilized text-davinci-002, given that the ReAct project\footnote{https://github.com/ysymyth/ReAct} has not incorporated gpt-3.5-turbo-1106. For the retrieval model, we used the Google Search API provided by SerpApi.com, following the ``Self-ask'' approach \citep{press2022measuring}. HGOT\footnote{https://github.com/fangyihao/hgot} was implemented using Python language and the DSP framework \citep{khattab2022demonstrate}. Following \citet{gao2023enabling}, We adopt a natural language inference (NLI) model \citep{honovich-etal-2022-true-evaluating} in HGOT to measure thought quality and retrieval quality. Additionally, the topological sorting and deductions pertaining to HGOT were performed using the Python NetworkX\footnote{https://networkx.org/} package. 

\renewcommand{\arraystretch}{1}
\begin{table*}[h]
\small
\centering

\setlength\tabcolsep{5pt}
\begin{tabular}{l|r|rr|rr||r|rr|rr}
\thickhline
\bf \multirow{2}{0.06\linewidth}{Method}&\multicolumn{1}{c|}{\bf FEVER}&\multicolumn{2}{c|}{\bf Open-SQuAD}&\multicolumn{2}{c||}{\bf HotPotQA}&\multicolumn{1}{c|}{\bf FEVER}&\multicolumn{2}{c|}{\bf Open-SQuAD}&\multicolumn{2}{c}{\bf HotPotQA}\\
\cline{2-11}
&\bf EM&\bf EM&\bf F1&\bf EM&\bf F1&\bf EM&\bf EM&\bf F1&\bf EM&\bf F1\\
\hline
&\multicolumn{5}{c||}{\bf Overall}&\multicolumn{5}{c}{\bf Long}\\
\cline{2-11}
Vanilla LM&54.72&17.43&33.91&33.58&43.93&43.36&16.10&34.22&24.09&38.15\\
Retrieve-then-Read&58.35&22.51&\bf 38.81&41.20&51.21&46.90&\bf 29.66&44.60&35.77&50.05\\
Self-ask&53.03&18.81&34.15&43.98&54.67&46.90&20.34&35.10&42.34&59.32\\
ReAct&45.04&-&-&35.47&42.18&34.51&-&-&17.52&24.62\\
DSP&55.45&20.65&36.09&47.23&\bf 61.13&47.79&23.73&39.08&\bf 45.26&\bf 64.27\\
\hline
HGOT+Sampling (Ours)&\bf 61.50&22.05&36.11&45.03&56.07&53.98&28.81&42.21&37.23&53.36\\
HGOT+KNN (Ours)&60.53&\bf 24.10&38.32&\bf 47.37&59.48&\bf 54.87&28.81&\bf 46.27&43.07&59.77\\
\hline\hline
&\multicolumn{5}{c||}{\bf Medium}&\multicolumn{5}{c}{\bf Short}\\
\cline{2-11}
Vanilla LM&54.00&26.42&41.10&29.73&40.63&64.00&9.43&26.49&44.59&51.59\\
Retrieve-then-Read&59.33&28.30&\bf 43.14&35.81&45.43&66.00&11.32&\bf 30.12&50.68&57.88\\
Self-ask&52.00&27.04&41.05&41.89&51.92&58.67&9.43&26.53&47.30&53.92\\
ReAct&45.33&-&-&33.11&40.69&52.67&-&-&51.35&56.89\\
DSP&55.33&28.93&42.51&41.89&57.17&61.33&10.06&27.41&\bf 54.05&\bf 62.72\\
\hline
HGOT+Sampling (Ours)&57.33&27.67&40.25&41.89&53.33&\bf 71.33&11.32&27.38&\bf 54.05&60.87\\
HGOT+KNN (Ours)&\bf 61.33&\bf 31.45&42.17&\bf 46.62&\bf 59.21&64.00&\bf 13.21&28.47&51.35&59.54\\
\thickhline
\end{tabular}
\caption{A comparative analysis of Vanilla LM, Retrieve-then-Read, Self-ask, ReAct, \texttt{DSP}, and HGOT. The ``Overall'' section is derived by calculating the weighted average of metrics from the ``Long'', ``Medium'', and ``Short'' categories, using the number of examples in each category as weights.}
\label{tab:experimental_results}
\vspace{-3mm}
\end{table*}
\renewcommand{\arraystretch}{1}

\section{Experimental Results}
\paragraph{Findings and Analysis: }
The baseline models, referred to as ``Vanilla LM'', utilize few-shot in-context learning on ChatGPT without being augmented by retrieval models. These ``Vanilla LM'' models closely mirror the fundamental capabilities of ChatGPT as assessed in our factuality evaluation datasets. We observe that ``Vanilla LM'' generally excels at responding to short questions (or claims in FEVER), except when it comes to short Open-SQuAD questions (refer to Table~\ref{tab:experimental_results}). This exception is consistent with our dataset analysis (see Appendix~\ref{sec:dataset_examples_and_examination} for details), where it is found that longer questions (or claims in FEVER) often demand the gathering of more facts and the undertaking of more complex reasoning. Conversely, questions of medium and short length in Open-SQuAD usually require identifying one or two specific pieces of knowledge. However, medium-length questions provide more context than the shorter ones.

Methods other than ``Vanilla LM'' include those that are augmented by retrieval mechanisms. In comparison, these retrieval-augmented approaches generally surpass the performance of ``Vanilla LM'', except in cases involving Self-ask and ReAct within the FEVER dataset (see the ``Overall'' section in Table~\ref{tab:experimental_results}). Additionally, the DSP method shows weaker performance in the FEVER dataset. This suggests that the ability to gather factual information is more crucial in FEVER than the capability for multi-hop reasoning. Our approaches, HGOT+Sampling and HGOT+KNN (with HGOT+Sampling and HGOT+KNN representing HGOT combined with the demonstration selection methods of ``balanced sampling'' or ``k-nearest neighbors'', as detailed in Appendix~\ref{sec:automated_annotated_demonstrations}), are versatile and exhibit strong performance across all three datasets, regardless of whether they prioritize the skill of accumulating factual data or conducting multi-hop comprehension and reasoning.

\begin{figure*}[h]
    \centering
        \includegraphics[width=0.915\linewidth]{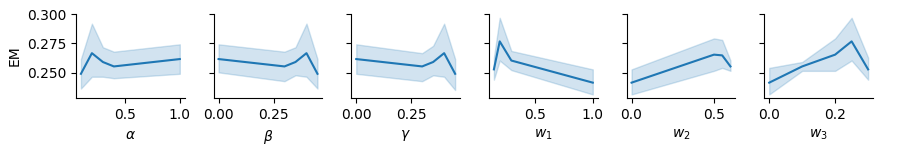}

    \caption{The visualizations of the hyperparameter searches are shown through pairwise relationships, featuring the EM score in the row and hyperparameters $\alpha$, $\beta$, $\gamma$, $w_1$, $w_2$, and $w_3$ in the columns. Each subplot is represented as a line chart, aggregating the data to display the mean (solid blue line) and the $95\%$ confidence interval (light blue area). Additionally, the optimal hyperparameters for attaining the highest EM score are indicated in each subplot.}
    \label{fig:grid_search}
\end{figure*}

Specifically for the FEVER dataset, HGOT+Sampling secures the top position, with HGOT+KNN closely behind in second place. With a $61.50\%$ EM score, HGOT+Sampling outperforms Retrieve-then-Read, which is third, by a margin of over $3\%$ (refer to the ``Overall'' section in Table~\ref{tab:experimental_results}). In every length category of the FEVER dataset, namely ``Long'', ``Medium'', and ``Short'', either HGOT+Sampling or HGOT+KNN achieves the highest ranking. Notably, HGOT+Sampling exceeds DSP, the strongest baseline, by more than $7\%$ in the ``Long'' category and surpasses Retrieve-then-Read by more than $5\%$ in the ``Short'' category, where Retrieve-then-Read is the top among baselines. In the ``Medium'' category, Retrieve-then-Read competes closely with HGOT+KNN, underscoring the importance of fact-gathering over complex reasoning in FEVER, in line with findings in Appendix~\ref{sec:dataset_examples_and_examination}. 
Moreover, both HGOT+Sampling and HGOT+KNN, on average, excel beyond Retrieve-then-Read's achievements in these scenarios.

Within the Open-SQuAD dataset, as detailed in Table~\ref{tab:experimental_results}'s ``Overall'' section, HGOT+KNN stands out as the top performer, recording an EM score of $24.10\%$, which is over $1.5\%$ higher than its nearest competitor, Retrieve-then-Read. HGOT+KNN also leads in EM scores for both the ``Medium'' and ``Short'' categories and achieves the highest F1 score in the ''Long'' category of the dataset. Retrieve-then-Read demonstrates strong competitiveness in the Open-SQuAD dataset, closely matching HGOT+KNN's performance across all categories, in contrast to DSP, which shows weaker performance. This observation is consistent with our analysis in Appendix~\ref{sec:dataset_examples_and_examination}, revealing that a large portion of the Open-SQuAD questions are designed to extract factual information, mainly asking ``What'', ``How'', and ``When''.

In the HotPotQA dataset, known for demanding multi-hop reasoning capabilities from models, HGOT+KNN achieved the top position in the total EM score. For the ``Medium'' category, HGOT+KNN recorded the highest EM score at $46.62\%$, surpassing the second-best performers, HGOT+Sampling, DSP, and Self-ask, by $4.73\%$. Additionally, in this category, HGOT+KNN led in F1 score, outperforming the second-ranked DSP by over $2\%$. DSP proved to be a strong contender across the board in the HotPotQA dataset, closely matching the performance of our HGOT+KNN model, whereas the Retrieve-then-Read model fell short. This performance trend corroborates our dataset examination in Appendix~\ref{sec:dataset_examples_and_examination}, confirming the necessity for models to possess robust multi-hop reasoning skills for the HotPotQA dataset.

\paragraph{Ablation Study: }
We examine the effect of the presence or absence of thought quality and retrieval quality, as well as how HGOT's performance varies with different hyperparameters. More precisely, we explore how the EM score interacts with the hyperparameters $\alpha$, $\beta$, and $\gamma$ as shown in Equation~\ref{eq:thought_quality}, and also how EM score relates to each element of $\vec{w}=(w_1, w_2, w_3)$ as detailed in Equation~\ref{eq:retrieval_quality}. Specifically, setting $\alpha=1$, $\beta=0$, and $\gamma=0$ in Equation~\ref{eq:thought_quality} is equivalent to a situation where thought quality is not considered, reducing the model to rely solely on prediction and calibration through self-consistency, as discussed in \citet{wang2023selfconsistency}. Similarly, when $w_1=1$, $w_2=0$, and $w_3=0$ in Equation~\ref{eq:retrieval_quality}, it simulates a condition where retrieval quality is disregarded, with the ranking of retrieved passages depending only on the search engine's score.

 We include hyperparameter settings of $\alpha=1$, $\beta=0$, and $\gamma=0$, alongside $w_1=1$, $w_2=0$, and $w_3=0$, to equalize the absence of thought quality and to simulate the absence of retrieval quality when searching for HGOT+KNN's optimal hyperparameter configurations for the medium-length category in the Open-SQuAD dataset. Figure~\ref{fig:grid_search} illustrates the EM scores associated with varying values of each hyperparameter. It is observed that the optimal EM score is attained with hyperparameter values of $\alpha=0.2$, $\beta=0.4$, $\gamma=0.4$, $w_1=0.2$, $w_2=0.55$, and $w_3=0.25$, as detailed in Table~\ref{tab:open-squad_grid_search} in Appendix~\ref{sec:extended_ablation_study}. This suggests that the optimal combination of hyperparameters can be identified with the presence of thought quality and retrieval quality, emphasizing the significance of introducing these qualities into the model (see Appendix~\ref{sec:extended_ablation_study} for additional results from the ablation study).

\section{Conclusion}

In our factuality evaluation, we chose FEVER, Open-SQuAD, and HotPotQA to assess models' abilities in both fact retrieval and reasoning. We segmented the datasets FEVER, Open-SQuAD, and HotPotQA into three categories: ``Long'', ``Medium'', and ``Short'', based on the length of their questions. This categorization emphasizes the significance of examining both extremely short and long questions, an aspect often overlooked in research. We introduced HGOT. This approach structures thoughts in a hierarchical graph format, leveraging emergent planning capabilities. It evaluates thoughts and retrieved passages by introducing metrics for thought and retrieval qualities, thereby safeguarding HGOT's capabilities in reasoning and fact-finding. Experiments show that HGOT stands out as a versatile approach, surpassing other models in FEVER and matching leading models such as Retrieve-then-Read in Open-SQuAD, and DSP in HotPotQA. 

\section*{Limitations}

HGOT employs OpenAI's ChatGPT for its language model, whereas alternative models such as Google's Gemini and Meta's Llama 2 have not been explored. HGOT's evaluation is conducted using the Google Search API from SerpApi.com as its retrieval model. Its performance could vary, either improve or decline, when used in conjunction with other search engines such as Microsoft Bing, Yahoo, and Baidu. Additionally, the retrieval model for HGOT could potentially include various domain-specific data sources, for example, this could involve aligning queries with pertinent information in relational databases such as Oracle and IBM's DB2, which are widely used in the finance industry. However, the effectiveness of these variant implementations has not been examined.

\section*{Ethics Statement}

We ensure that all data utilized is publicly available and refrain from involving any private data. We affirm that our research focuses on assessing factuality and deliberately avoids producing harmful or undesirable content. 

\bibliography{custom}

\appendix

\newpage
\onecolumn

\section{Dataset Summary Statistics}
\label{sec:dataset_summary_statistics}

Table~\ref{tab:summary_statistics} presents a comparison of the FEVER, Open-SQuAD, and HotPotQA datasets across nine evaluated categories in our experiments. For each category, we assess the total number of instances, as well as the maximum, minimum, and median lengths of questions, in addition to calculating the mean and standard deviation for question lengths. It is noted that the question lengths in all three categories of the Open-SQuAD dataset are generally shorter compared to the equivalent categories in the FEVER and HotPotQA datasets. Furthermore, the ``Long'' and ``Medium'' categories exhibit larger standard deviations in question length across all three datasets when compared to the ``Short'' categories.

\renewcommand{\arraystretch}{1}
\begin{table}[h]
\small
\centering
\begin{tabular}{p{0.07\linewidth}|p{0.08\linewidth}|p{0.06\linewidth}|p{0.11\linewidth}p{0.09\linewidth}p{0.09\linewidth}p{0.07\linewidth}p{0.07\linewidth}p{0.1\linewidth}}
\thickhline

\bf Dataset & \bf Sentence Length & \bf Split &\bf Number of Examples&\bf Maximum Length&\bf Minimum Length&\bf Median&\bf Mean&\bf Standard Deviation\\
\hline
\multirow{9}{*}{\rotatebox{90}{FEVER}}&\multirow{3}{*}{Long}&Train&1619&125&38&40&44.33&12.89\\
&&Dev&113&57&37&38&39.22&3.58\\
&&Test&113&53&39&41&42.33&3.24\\
\cline{2-9}
&\multirow{3}{*}{Medium}&Train&2182&37&24&27&27.51&2.82\\
&&Dev&150&36&24&27&27.49&2.63\\
&&Test&150&37&24&27&27.81&2.90\\
\cline{2-9}
&\multirow{3}{*}{Short}&Train&2182&23&21&23&22.81&0.40\\
&&Dev&150&23&21&23&22.81&0.41\\
&&Test&150&23&22&23&22.76&0.43\\

\hline
\hline
\multirow{9}{*}{\rotatebox{90}{Open-SQuAD}}&\multirow{3}{*}{Long}&Train&1174&60&22&23&24.42&3.18\\
&&Dev&121&36&22&24&24.55&2.86\\
&&Test&118&34&23&24&25.02&2.55\\
\cline{2-9}
&\multirow{3}{*}{Medium}&Train&1181&21&6&11&11.26&3.29\\
&&Dev&133&20&6&11&11.41&3.29\\
&&Test&159&19&6&11&11.53&3.34\\
\cline{2-9}
&\multirow{3}{*}{Short}&Train&1181&5&1&5&4.72&0.57\\
&&Dev&133&5&4&5&4.83&0.38\\
&&Test&159&5&3&5&4.79&0.47\\

\hline
\hline
\multirow{9}{*}{\rotatebox{90}{HotPotQA}}&\multirow{3}{*}{Long}&Train&1504&128&58&66&69.46&10.96\\
&&Dev&168&120&59&66&69.12&10.31\\
&&Test&137&57&34&36&37.66&3.98\\
\cline{2-9}
&\multirow{3}{*}{Medium}&Train&1628&57&10&17&19.49&8.33\\
&&Dev&181&58&10&18&20.23&9.80\\
&&Test&148&33&10&17&17.71&5.43\\
\cline{2-9}
&\multirow{3}{*}{Short}&Train&1628&9&4&9&8.43&0.91\\
&&Dev&181&9&5&9&8.43&0.90\\
&&Test&148&9&7&9&8.57&0.65\\
\thickhline
\end{tabular}

\caption{Summary statistics across three datasets FEVER, Open-SQuAD, and HotPotQA and nine categories}
\label{tab:summary_statistics}
\end{table}
\renewcommand{\arraystretch}{1}

\section{Dataset Examples and Examination}
\label{sec:dataset_examples_and_examination}
\renewcommand{\arraystretch}{1.45}

\subsection{FEVER Data Examples and Examination}

The FEVER dataset necessitates that the model gathers relevant background information or context regarding the subject, such as knowing what the Boeing 767 is as stated in the claim ``The Boeing 767 became the most frequently used airliner for transatlantic flights between North America and Europe in the 1990s'' (Table~\ref{tab:fever_data_examples}). Subsequently, it is required to conduct logical analysis on all the specific facts collected. Claims that are longer typically require the accumulation of more facts and knowledge, as well as the undertaking of more sophisticated reasoning. As a result, the complexity of a claim is often proportional to its length.

\begin{table}[H]
\centering
\begin{tabular}{p{0.1\linewidth}|p{0.57\linewidth}|p{0.23\linewidth}}
\thickhline
\bf Sentence Length&\bf Claim&\bf Answer\\
\hline
\multirow{5}{*}{Long}

&The Boeing 767 became the most frequently used airliner for transatlantic flights between North America and Europe in the 1990s.&SUPPORTS\\

&In Kentucky, the electric chair has been kept in operation except for those whose capital crimes were committed prior to March 31, 1998, and who choose electrocution.&REFUTES\\

&The House of the Spirits is about the life of a young lady named Clara during the military dictatorship in Algeria.&REFUTES\\

&One Flew Over the Cuckoo's Nest won the five major Academy Awards the year it was released, the second film to do so.&NOT ENOUGH INFO\\

&In 2012, Simi Valley, California, reported a higher median household income than that of the nation overall.&SUPPORTS\\

\hline\hline
\multirow{5}{*}{Medium}

&Planet Hollywood Las Vegas is operated by all entities except an American gaming corporation.&REFUTES\\

&Chris Bosh plays in the National Basketball Association as a professional basketball player.&SUPPORTS\\

&Pierce County, Washington is the location of the lowest mountain in Washington.&NOT ENOUGH INFO\\

&The Airbus A380 entered commercial service on October 25, 2017.&REFUTES\\

&The Nobel Prize in Chemistry was awarded to a person from the Kingdom of the Netherlands.&SUPPORTS\\

\hline\hline
\multirow{5}{*}{Short}

&Estonia is a country.&SUPPORTS\\

&Edward Cullen was created.&NOT ENOUGH INFO\\

&Dopamine prevents neuromodulation.&REFUTES\\

&Backing vocalists are performers.&SUPPORTS\\

&Reanimation is a book.&NOT ENOUGH INFO\\

\thickhline
\end{tabular}
\caption{FEVER data examples}
\label{tab:fever_data_examples}
\end{table}

\subsection{Open-SQuAD Data Examples and Examination}

As demonstrated in Table~\ref{tab:open-squad_data_examples} of the Open-SQuAD dataset, the bulk of questions are focused on ``What'', ``How'', ``When'', and ``Why'', requiring the accumulation of factual data for answers. Additionally, questions of medium and short length typically need the collection of one or two specific pieces of information or knowledge. For instance, the question ``In what geographical portion of Wales is Abercynon located?'' necessitates identifying the specific location of Abercynon within Wales. Notably, medium-length questions tend to offer more context for information retrieval compared to those in the ``Short'' category, such as ``What is septicemia?''. Thus, the inclusion of ``Short'' category questions in Open-SQuAD doesn't suggest they are easy to answer, especially for models that find it challenging to gather factual data. Conversely, ``Long'' category questions usually demand more extensive fact-finding and complex reasoning.

\begin{table}[H]
\centering
\begin{tabular}{p{0.1\linewidth}|p{0.57\linewidth}|p{0.23\linewidth}}
\thickhline
\bf Sentence Length&\bf Question&\bf Answer\\
\hline
\multirow{5}{*}{Long}

&What was the number of times the Denver Broncos played in a Super Bowl by the time they reached Super Bowl 50?&eight\\

&What is the application of prime numbers used in information technology which utilizes the fact that factoring very large prime numbers is very challenging?&public-key cryptography\\

&When did the UMC's General Board of Church and Society call on all United Methodists to abstain from alcohol for Lent?&2011 and 2012\\

&What is the minimum distance between a patient's home and the nearest pharmacy that allows a physician in Austria to give out medicine?&more than 4 kilometers\\

&Approximately how many names were signed on an online petition on the Parliamentary website in response to the closing of the Musical Instruments gallery?&over 5,100\\

\hline\hline
\multirow{5}{*}{Medium}

&In what geographical portion of Wales is Abercynon located?&south\\

&How long has the Doctor Who Magazine been in circulation?&since 1979\\

&What social construct did Huguenot refugees in Canterbury practice?&economic separation\\

&Why were Johann Esch and Heinrich Voes executed by the Catholic Church?&for Lutheran views\\

&Who was the first known European to visit China and return?&Marco Polo\\

\hline\hline
\multirow{5}{*}{Short}

&What is septicemia?&a type of ``blood poisoning''\\

&What shape are Plastoglobuli?&spherical bubbles\\

&What do carotenoids absorb?&light energy\\

&What is a prasinophyte?&a green algal derived chloroplast\\

&What was Apple Talk&a proprietary suite of networking protocols developed by Apple Inc\\

\thickhline
\end{tabular}
\caption{Open-SQuAD data examples}
\label{tab:open-squad_data_examples}
\end{table}

\subsection{HotPotQA Data Examples and Examination}

HotPotQA questions typically demand from the model not only the skill to accumulate factual data but, more importantly, the capability for multi-hop comprehension and reasoning, particularly with long questions. For instance, to answer the question (refer to Table~\ref{tab:hotpotqa_data_examples}), ``What is the genus of the viral disease that has symptoms such as fever, chills, loss of appetite, nausea, muscle pains, and headaches, and has a chance of causing liver damage?'' the model is required to initially identify details about ``the viral disease that has symptoms such as fever, chills, loss of appetite, nausea, muscle pains, and headaches'' alongside information on ``the viral disease that has a chance of causing liver damage'', before determining the genus of the virus in question. Therefore, the degree of complexity for a HotPotQA question often correlates with its length.

\begin{table}[H]
\centering

\begin{tabular}{p{0.1\linewidth}|p{0.57\linewidth}|p{0.23\linewidth}}
\thickhline
\bf Sentence Length&\bf Question&\bf Answer\\
\hline
\multirow{5}{*}{Long}

&Out of two American colonies that had a series of skirmishes and raids between 1701 and 1765 at the disputed border, which British proprietary colony became a royal colony on the northeast coast of North America?&Province of New York\\

&Which Captain launched the attack which led to more casualties than any other incident in the war fought between the settlers of the nascent colony of New Netherland and the native Lenape population?&Captain John Underhill\\

&Lost Kingdom Adventure is a dark ride located at four Legoland theme parks, including which park, which is the original Legoland park, that was opened on June 7th, 1968?&Legoland Billund\\

&What is the genus of the viral disease that has symptoms such as fever, chills, loss of appetite, nausea, muscle pains, and headaches, and has a chance of causing liver damage?&Flavivirus\\

&Until what year did the Chief of Justice of the Supreme Court that administered the presidential oath of office to Abraham Lincoln on his first inauguration as the 16th President of the United States hold that office?&1864\\

\hline\hline
\multirow{5}{*}{Medium}

&The Last Run is a drama film that stars which Lithuanian-American actor?&Vyto Ruginis\\

&What part of Australia is Alice River and Rupertswood in?&Victoria\\

&What was the nationality of the composer of Chaconne in F minor?&German\\

&What was the breakthrough role of the actor starring in Good Boy! and was a native of Atlanta?&Tai Frasier in ``Clueless''\\

&Who played the role of Nettie Harris in the 1985 film directed by Steven Spielberg?&Akosua Gyamama Busia\\

\hline\hline
\multirow{5}{*}{Short}

&What empire was Aleksei Gen born into?&Russian Empire\\

&Romans stars which Tamil and Telugu actress?&Nivetha Thomas\\

&Are Ari Up and Boz Burrell both guitarists?&no\\

&Are Tetrastigma and Spruce both types of plants?&yes\\

&What did Karan Kapoor's maternal grandfather deliver?&Shakespeare performances\\

\thickhline
\end{tabular}
\caption{HotPotQA data examples}
\label{tab:hotpotqa_data_examples}
\end{table}

\renewcommand{\arraystretch}{1}

\newpage
\section{Prompt and Response Examples}
\subsection{Prompt and Response of the ``Plan'' Procedure}
\label{sec:plan_prompt_and_response}
\begin{lstlisting}[]
~~~~~~~~~~~~~~~~~~~~~~~~~~~~~~~~~~~ PROMPT ~~~~~~~~~~~~~~~~~~~~~~~~~~~~~~~~~~~
...................................user...................................
Sketch a plan to answer the following question with the provided context. List only the essential steps which can be answered by search engines. Express each step as a standalone search question. Highlight interdependencies if any. Higher number steps can depend on lower number steps, while the reverse is not possible.

---

Follow the following format.

Context:
${sources that may contain relevant content. e.g., [1] Passage 1. [2] Passage 2. [3] Passage 3.}

Question: ${the question to be answered}

Plan:
Step 1: ${a standalone search question. e.g., ...?} Step 2: ${a standalone search question. e.g., ...?} ... Step n: ${a standalone search question. e.g., ...?}

Dependencies: ${interdependencies among multiple steps. e.g., Step ... depends on Step ... .}

---

Context:
[1] Steve Masiello | (born September 2, 1977) is an American college basketball coach and a former player. He most recently served as men's head coach at Manhattan College.
[2] Jaspers' new coach hopes to recapture MC's past glory | Manhattan College introduced Steve Masiello, center, who will take over as the Jaspers' new men's basketball coach.
[3] Steve Masiello (St. John's Red Storm) | Steve Masiello (born September 2, 1977). Current position: Associate head men's basketball coach. Current team: St. John's Red Storm (Head ...

Question: Which of the Manhattan Jaspers basketball team head coach was born in September 2, 1977?

Plan:
Step 1: Who is the head coach of the Manhattan Jaspers basketball team? Step 2: When was the head coach born?

Dependencies: Step 2 depends on Step 1.

---

Context:
[1] Phil Cutchin | Phil Cutchin (September 9, 1920 - January 7, 1999) was an American football player and coach. He served as the head football coach at Oklahoma State ...
[2] Former OSU Football Coach Cutchin Dies | In life, Phil Cutchin captained a Paul "Bear" Bryant football team, was an Army officer in two wars, a football coach and a stock broker.
[3] Phil Cutchin | American Football Database | Fandom | Phil Cutchin (September 9, 1920 - January 7, 1999) was an American football player and coach. He served as the head football coach at Oklahoma State ...

Question: Coach Phil Cutchin served as the head football coach at Oklahoma State-University-Stillwater, which was originally known as what?

Plan:
Step 1: What was Oklahoma State University-Stillwater originally known as? Step 2: When did Phil Cutchin serve as the head football coach at Oklahoma State University-Stillwater?

Dependencies: Step 2 depends on Step 1.

---

Context:
Todd Boehly | Todd Boehly is an American businessman and investor. He is the co-founder, chairman, chief executive officer and controlling member of Eldridge Industries, ...

Question: What was Todd Boehly's former position at the firm where Mark Walter is the CEO?

Plan:
----------------------------------- RESPONSE -----------------------------------
----------------------------------- CHOICE 0 -----------------------------------
...................................assistant...................................
Step 1: What is the name of the firm where Mark Walter is the CEO? Step 2: What was Todd Boehly's former position at the firm where Mark Walter is the CEO?

Dependencies: Step 2 depends on Step 1.
\end{lstlisting}

\subsection{Prompt and Response of the ``Self-reflect'' Procedure}
\label{sec:self_reflect_prompt_and_response}
\begin{lstlisting}[]
~~~~~~~~~~~~~~~~~~~~~~~~~~~~~~~~~~~ PROMPT ~~~~~~~~~~~~~~~~~~~~~~~~~~~~~~~~~~~
...................................user...................................
Highlight interdependencies among the steps below if any. Higher number steps can depend on lower number steps, while the reverse is not possible.

---

Follow the following format.

Plan:
Step 1: ${a standalone search question. e.g., ...?} Step 2: ${a standalone search question. e.g., ...?} ... Step n: ${a standalone search question. e.g., ...?}

Dependencies: ${interdependencies among multiple steps. e.g., Step ... depends on Step ... .}

---

Plan:
Step 1: Who is the head coach of the Manhattan Jaspers basketball team? Step 2: When was the head coach born?

Dependencies: Step 2 depends on Step 1.

---

Plan:
Step 1: What was Oklahoma State University-Stillwater originally known as? Step 2: When did Phil Cutchin serve as the head football coach at Oklahoma State University-Stillwater?

Dependencies: Step 2 depends on Step 1.

---

Plan:
Step 1: What is the name of the firm where Mark Walter is the CEO? Step 2: What was Todd Boehly's former position at the firm where Mark Walter is the CEO?

Dependencies:
----------------------------------- RESPONSE -----------------------------------
----------------------------------- CHOICE 0 -----------------------------------
...................................assistant...................................
Step 2 depends on Step 1.
\end{lstlisting}

\subsection{Prompt and Response of the ``Formalize'' Procedure}
\label{sec:formalize_prompt_and_response}
\begin{lstlisting}[]
~~~~~~~~~~~~~~~~~~~~~~~~~~~~~~~~~~~ PROMPT ~~~~~~~~~~~~~~~~~~~~~~~~~~~~~~~~~~~
...................................user...................................
Express the dependencies in formal language by giving the descriptions below.

---

Follow the following format.

Descriptions: ${descriptions of dependencies}
Dependencies: ${e.g., If Step 2 depends on Step 1, then write Step 1 -> Step 2; If Step 2 and Step 3 depend on Step 1, then write Step 1 -> (Step 2 and Step 3); If Step 3 depends on Step 1 and Step 2, then write (Step 1 and Step 2) -> Step 3}

---

Descriptions: Step 2 depends on Step 1.
Dependencies:
----------------------------------- RESPONSE -----------------------------------
----------------------------------- CHOICE 0 -----------------------------------
...................................assistant...................................
Step 1 -> Step 2
\end{lstlisting}
\subsection{Prompt and Response of the ``Rewrite'' Procedure}
\label{sec:rewrite_prompt_and_response}
\begin{lstlisting}[]
~~~~~~~~~~~~~~~~~~~~~~~~~~~~~~~~~~~ PROMPT ~~~~~~~~~~~~~~~~~~~~~~~~~~~~~~~~~~~
...................................user...................................
Rewrite the last question in a standalone manner by giving the answers to previous questions. Do not consider answers that were not specified. Only show the last question after the rewrite.

---

Follow the following format.

Context:
${previous questions and answers}

Rewrite: ${the last question after the rewrite}

---

Context:
Step 1: Who is the head coach of the Manhattan Jaspers basketball team? ANSWER: John Gallagher. Step 2: When was the head coach born?

Rewrite: When was the head coach of the Manhattan Jaspers basketball team born?

---

Context:
Step 1: What was Oklahoma State University-Stillwater originally known as? ANSWER: Oklahoma Agricultural and Mechanical College. Step 2: When did Phil Cutchin serve as the head football coach at Oklahoma State University-Stillwater?

Rewrite: When did Phil Cutchin serve as the head football coach at Oklahoma State University-Stillwater?

---

Context:
Step 1: What is the name of the firm where Mark Walter is the CEO? ANSWER: Guggenheim Partners. Step 2: What was Todd Boehly's former position at the firm where Mark Walter is the CEO?

Rewrite:
----------------------------------- RESPONSE -----------------------------------
----------------------------------- CHOICE 0 -----------------------------------
...................................assistant...................................
What was Todd Boehly's former position at Guggenheim Partners?
\end{lstlisting}
\subsection{Prompt and Response of the ``Predict'' Procedure}
\label{sec:predict_prompt_and_response}
\begin{lstlisting}[]
~~~~~~~~~~~~~~~~~~~~~~~~~~~~~~~~~~~ PROMPT ~~~~~~~~~~~~~~~~~~~~~~~~~~~~~~~~~~~
...................................user...................................
Answer questions with short factoid answers.

---

Follow the following format.

Context:
${sources that may contain relevant content. e.g., [1] Passage 1. [2] Passage 2. [3] Passage 3.}

Question: ${the question to be answered}

Rationale: Let's think step by step. ${a step-by-step deduction that identifies the correct response, which will be provided below. Every statement in the "Rationale" section should be attributable to the passages provided in the "Context" section. e.g., ...[1][2].}

Answer: ${a short factoid answer, often between 1 and 5 words}

---

Context:
[1] List of Manhattan Jaspers men's basketball head coaches | Manhattan's current head coach is John Gallagher. He was hired in March 2023, replacing RaShawn Stores, who was not promoted to the full-time position after ...
[2] Steve Masiello | Stephen John Masiello Jr. (born September 2, 1977) is an American college basketball coach and a former player. He most recently served as men's head coach ...
[3] Steve Masiello | (born September 2, 1977) is an American college basketball coach and a former player. He most recently served as men's head coach at Manhattan College.
[4] Manhattan College Appoints John Gallagher to Lead Men's ... | - John Gallagher has been named the new Head Men's Basketball Coach at Manhattan College, it was announced today by Director of Athletics ...
[5] List of Manhattan Jaspers men's basketball head coaches | Manhattan's current head coach is John Gallagher. He was hired in March 2023, replacing RaShawn Stores, who was not promoted to the full-time position after ...
[6] Jaspers' new coach hopes to recapture MC's past glory | Manhattan College introduced Steve Masiello, center, who will take over as the Jaspers' new men's basketball coach.
[7] Men's Basketball Coaches | Head Coach, 718-862-7533 718-862-7533 . jgallagher06@manhattan.edu, First Year ; Assistant Coach, 718-862-7533 718-862-7533 . tim.brooks@manhattan.edu, First ...

Question: Which of the Manhattan Jaspers basketball team head coach was born in September 2, 1977?

Rationale: Let's think step by step. Steve Masiello was born on September 2, 1977 [2][3]. John Gallagher is the current head coach of the Manhattan Jaspers basketball team [1][4][5].

Answer: Steve Masiello

---

Context:
[1] Oklahoma Agricultural and Mechanical College | Oklahoma Agricultural and Mechanical College, Founded on Christmas Day in 1890 under the Morrill Act as Oklahoma Agricultural and Mechanical College, Oklahoma State University has grown through its traditions and culture to become one of America's premier land-grant universities., Oklahoma Agricultural and Mechanical College
[2] Oklahoma State University-Stillwater | OSU was founded in 1890 under the Morrill Act. Originally known as Oklahoma Agricultural and Mechanical College (Oklahoma A&M), it is the flagship institution ...
[3] 1963 to 1968 | 1963 to 1968, Phil Cutchin (September 9, 1920 - January 7, 1999) was an American football player and coach. He served as the head football coach at Oklahoma State University-Stillwater from 1963 to 1968, compiling a record of 19-38-2., 1963 to 1968
[4] Former OSU Football Coach Cutchin Dies | Cutchin was head football coach at Oklahoma State from 1963 to 1968. He won only 19 games, but most all of his 40 defeats were given up ...
[5] Phil Cutchin | Phil Cutchin (September 9, 1920 - January 7, 1999) was an American football player and coach. He served as the head football coach at Oklahoma State ...
[6] OSU History | The college's first students attended classes in the Stillwater Congregational Church. The original campus consisted of 200 acres of prairie that were ...
[7] Phil Cutchin | American Football Database | Fandom | He served as the head football coach at Oklahoma State University-Stillwater from 1963 to 1968, compiling a record of 19-38-2. Although he never had a winning ...

Question: Coach Phil Cutchin served as the head football coach at Oklahoma State-University-Stillwater, which was originally known as what?

Rationale: Let's think step by step. Oklahoma Agricultural and Mechanical College [1][2].

Answer: Oklahoma Agricultural and Mechanical College

---

Context:
[1] Unions file lawsuit challenging Wisconsin Act 10 | Former Republican Gov. Scott Walker signed the law in 2011 despite some of the largest protests in state history, and the law has since shaped the state's political landscape., Scott Walker
[2] Act 10 turns 10: Four takeaways from the law that shook ... | Here's a look at how the law limiting collective bargaining for most public workers has played out.
[3] Act 10 turns 10: Four takeaways from the law that shook ... | Act 10 ended the ability of public-sector unions to negotiate over any issues other than raises, and those raises were capped at the rate of ...
[4] Wisconsin Teachers Sue to Restore Collective Bargaining ... | The law, which was championed by former Republican Gov. Scott Walker, has been challenged unsuccessfully in court before. But the political context has changed: The Wisconsin Supreme Court recently flipped to liberal control for the first time in 15 years., Scott Walker
[5] Wis. governor officially cuts collective bargaining | Scott Walker has officially taken away nearly all collective bargaining rights from the vast majority of the state's public employees. Walker ...
[6] 10 years later, Wisconsinites are still divided over Act 10 | Former Gov. Scott Walker's landmark legislation required public employees to pay more for their pensions and health care and limited their ...
[7] Wisconsin's Act 10 limitations on collective bargaining | With its 5-2 vote upholding the law, the Wisconsin Supreme Court gave an important nod towards the constitutionality of limits of collective bargaining rights ...

Question: Which Wisconsin state governor oversaw a vote to significantly limit public employee collective bargaining?

Rationale: Let's think step by step. Former Republican Governor Scott Walker oversaw a vote to significantly limit public employee collective bargaining [1][4][5][6][7].

Answer: Scott Walker

---

Context:
[1] Mark Walter | 184 Mark Walter on the 2023 Forbes 400 - Mark Walter is CEO of investment firm Guggenheim Partners, which has over $300 billion in assets under management.
[2] Todd Boehly - Milken Institute | Boehly was the President of Guggenheim Partners. He received his B.B.A. from the College of William & Mary, where he later founded the Boehly Center for Excellence in Finance, and studied at the London School of Economics., President
[3] Katie & Todd Boehly | Prior to founding Eldridge Industries, Todd was President of Guggenheim Partners and founded the credit business at Guggenheim. He received his B.B.A. from The College of William & Mary, where he later founded The Boehly Center for Excellence in Finance, and studied at the London School of Economics., Prior to founding Eldridge Industries, Todd was President of Guggenheim Partners and founded the credit business at Guggenheim
[4] Katie & Todd Boehly | Prior to founding Eldridge Industries, Todd was President of Guggenheim Partners and founded the credit business at Guggenheim., President of Guggenheim Partners
[5] Guggenheim Partners | Guggenheim Partners, In 2000, he helped found Guggenheim Partners. Today, he is the CEO of Guggenheim Partners, which has rapidly grown into a global, diversified financial services firm with more than $260 billion in assets under management, 2,300 employees, and 25 offices in six countries around the world., Guggenheim Partners
[6] Todd Boehly | Todd Boehly cofounded holding company Eldridge in 2015, after building Guggenheim Partners' credit business and serving as president. Boehly acquired some ...
[7] Todd L. Boehly | Boehly was President of Guggenheim Partners and founded the credit business at Guggenheim. He received his B.B.A. from The College of William & Mary, where ...

Question: What was Todd Boehly's former position at the firm where Mark Walter is the CEO?

Rationale: Let's think step by step.
----------------------------------- RESPONSE -----------------------------------
----------------------------------- CHOICE 0 -----------------------------------
...................................assistant...................................
Todd Boehly was the President of Guggenheim Partners [2][3][4][6][7].

Answer: President
----------------------------------- CHOICE 1 -----------------------------------
...................................assistant...................................
Todd Boehly was the President of Guggenheim Partners [2][3][4][6][7].

Answer: President
----------------------------------- CHOICE 2 -----------------------------------
...................................assistant...................................
Todd Boehly was the President of Guggenheim Partners [2][3][4][6][7].

Answer: President
----------------------------------- CHOICE 3 -----------------------------------
...................................assistant...................................
Todd Boehly was the President of Guggenheim Partners [2][3][4][6][7].

Answer: President
----------------------------------- CHOICE 4 -----------------------------------
...................................assistant...................................
Todd Boehly was the President of Guggenheim Partners [2][3][4][7].

Answer: President
----------------------------------- CHOICE 5 -----------------------------------
...................................assistant...................................
Todd Boehly was the President of Guggenheim Partners [2][3][4][7].

Answer: President
----------------------------------- CHOICE 6 -----------------------------------
...................................assistant...................................
Todd Boehly was the President of Guggenheim Partners [2][3][4][6][7].

Answer: President
----------------------------------- CHOICE 7 -----------------------------------
...................................assistant...................................
Todd Boehly was the President of Guggenheim Partners [2][3][4][6][7].

Answer: President
----------------------------------- CHOICE 8 -----------------------------------
...................................assistant...................................
Todd Boehly was the President of Guggenheim Partners [2][3][4][6][7].

Answer: President
----------------------------------- CHOICE 9 -----------------------------------
...................................assistant...................................
Todd Boehly was the President of Guggenheim Partners [2][3][4][6][7].

Answer: President
----------------------------------- CHOICE 10 -----------------------------------
...................................assistant...................................
Todd Boehly was the President of Guggenheim Partners [2][3][4][6][7].

Answer: President
----------------------------------- CHOICE 11 -----------------------------------
...................................assistant...................................
Todd Boehly was the President of Guggenheim Partners [2][3][4][7].

Answer: President
----------------------------------- CHOICE 12 -----------------------------------
...................................assistant...................................
Todd Boehly was the President of Guggenheim Partners [2][3][4][6][7].

Answer: President
----------------------------------- CHOICE 13 -----------------------------------
...................................assistant...................................
Todd Boehly was the President of Guggenheim Partners [2][3][4][6][7].

Answer: President
----------------------------------- CHOICE 14 -----------------------------------
...................................assistant...................................
Todd Boehly was the President of Guggenheim Partners [2][3][4][6][7].

Answer: President
----------------------------------- CHOICE 15 -----------------------------------
...................................assistant...................................
Todd Boehly was the President of Guggenheim Partners [2][3][4][6][7].

Answer: President
----------------------------------- CHOICE 16 -----------------------------------
...................................assistant...................................
Todd Boehly was the President of Guggenheim Partners [2][3][4][7].

Answer: President
----------------------------------- CHOICE 17 -----------------------------------
...................................assistant...................................
Todd Boehly was the President of Guggenheim Partners [2][3][4][6][7].

Answer: President
----------------------------------- CHOICE 18 -----------------------------------
...................................assistant...................................
Todd Boehly was the President of Guggenheim Partners [2][3][4][6][7].

Answer: President
----------------------------------- CHOICE 19 -----------------------------------
...................................assistant...................................
Todd Boehly was the President of Guggenheim Partners [2][3][4][6][7].

Answer: President
----------------------------------- ANSWER -----------------------------------
President
----------------------------------- CONFIDENCE -----------------------------------
1.0
\end{lstlisting}

\section{Automated Annotated Demonstrations}
\label{sec:automated_annotated_demonstrations}

Following DSP \cite{khattab2022demonstrate}, a demonstration is defined as a training example crafted to illustrate particular behaviors expected from the LLM. A qualifying example of such a demonstration occurs when the model's prediction for this example aligns with the actual correct answer. We extend DSP's approach by incorporating additional considerations into the automated creation of demonstrations. 

In the automated creation of demonstrations for use in the ``Probe'' and ``Infer'' procedures, we adjust citation marks using regular expressions. We employ the regular expression \bverb|(\[[0-9]+\])+| to identify citation marks and ensure they are placed at the end of each sentence or statement, if they are not already. To verify that all sentences or statements adhere to this format, we use the regular expression \bverb|^([^\[\.]+(\[[0-9]+\])*\.)+$|. This standardized format aids in accurately tallying the total count of cited passages.

For demonstrations intended for the ``Plan'' procedure, we select premium dependency rules utilizing regular expressions. The regular expression \bverb/None|((\s*([Ss]tep [0-9]+) depends on ([Ss]tep [0-9]+)\.\s*)+)/ is used to ensure that dependencies in the dependency graph, generated by LLM, conform to a particular format. This assists in the precise identification of these relationships.

During our observations in automated annotated demonstrations for the ``Plan'' procedure, we have noticed that overly long sub-queries or steps produced by LLM often erroneously repeat the original, more complex question, deviating from the divide-and-conquer strategy of breaking down a complex question into smaller sub-queries. To address this, we implement the outlier detection method known as the interquartile range (IQR) to identify and disqualify any excessively long sub-query or step.

In selecting demonstrations for a prompt, we utilize two different approaches: balanced sampling and k-nearest neighbors (KNN). Balanced sampling involves randomly selecting from training examples while making sure to maintain an even distribution of answers (classes). KNN, on the other hand, makes use of sentence representations\footnote{https://huggingface.co/sentence-transformers/all-MiniLM-L6-v2} to identify and select the k training examples closest to the input question (or claim, as in the case of FEVER). This approach was investigated by \citet{liu-etal-2022-makes}.

\section{Baselines}
\label{sec:baselines}

Our benchmarking encompasses five methods: ``Vanilla LM'' as outlined by \citet{brown2020language}, ``Retrieve-then-Read'' as discussed in the works of \citet{lazaridou2022internet} and \citet{izacard2022few}, ``Self-ask'' introduced by \citet{press2022measuring}, ``ReAct'' described by \citet{yao2023react}, and ``Demonstrate-Search-Predict'' (DSP) presented by \citet{khattab2022demonstrate}.

\begin{itemize}
\item
Vanilla LM: The ``Vanilla LM'' baselines employ the few-shot in-context learning approach as proposed by \citet{brown2020language}. These basic benchmarks don't engage in retrieving text passages pertinent to the input query.
\item
Retrieve-then-Read: The ``Retrieve-then-Read'' benchmarks utilize the retrieval model (RM) to support each instance with a possibly relevant text passage prior to presenting the prompt to the language model (LM). 
\item
Self-ask: The ``Self-ask'' baselines involve the LM posing additional ``follow-up questions'' that are then 
directed to a retrieval model. Adhering to \citet{khattab2022demonstrate}, we alter the Self-ask's prompt design by:
(i) merging few-shot training instances from the task, such as question-answer pairs, at the beginning of the prompt,
(ii) instructing the model to produce a brief initial answer at each retrieval phase, and
(iii) specifically commanding the model to generate a subsequent ``search query'' at each stage.
\item
ReAct: The ReAct method utilizes LLMs to concurrently create reasoning traces and task-specific actions. We test ReAct using the ``text-davinci-002'' backbone LLM, focusing on the FEVER and HotPotQA datasets. However, the ReAct project has not incorporated the Open-SQuAD dataset and the ``gpt-3.5-turbo-1106'' backbone LLM, thus these have not been subjected to evaluation.
\item
Demonstrate-Search-Predict (DSP): The DSP method initiates pipeline-aware demonstrations, seeks out related passages, and creates predictions rooted in evidence. Following \citet{khattab2022demonstrate}, we utilize random sampling to select and annotate examples, and then employ them as demonstrations.
\end{itemize}




\section{Extended Ablation Study}
\label{sec:extended_ablation_study}

\begin{table}[H]
\centering
\small
\begin{tabular}{p{0.1\linewidth}|p{0.1\linewidth}|p{0.1\linewidth}|p{0.1\linewidth}|p{0.1\linewidth}|p{0.1\linewidth}|p{0.1\linewidth}|p{0.1\linewidth}}
\thickhline
\bf $\alpha$&\bf $\beta$&\bf $\gamma$&\bf $w_1$&\bf $w_2$&\bf $w_3$&\bf EM&\bf F1\\
\hline
0.1&0.45&0.45&0.15&0.55&0.3&25.16&36.55\\
0.1&0.45&0.45&0.2&0.55&0.25&27.04&39.34\\
0.1&0.45&0.45&0.3&0.5&0.2&24.53&35.20\\
0.1&0.45&0.45&0.3&0.6&0.1&25.16&35.35\\
0.1&0.45&0.45&1&0&0&22.64&34.15\\
0.2&0.4&0.4&0.15&0.55&0.3&25.16&36.55\\
0.2&0.4&0.4&0.2&0.55&0.25&31.45&42.17\\
0.2&0.4&0.4&0.3&0.5&0.2&27.67&41.44\\
0.2&0.4&0.4&0.3&0.6&0.1&25.16&35.40\\
0.2&0.4&0.4&1&0&0&23.90&35.27\\
0.3&0.35&0.35&0.15&0.55&0.3&23.90&37.03\\
0.3&0.35&0.35&0.2&0.55&0.25&25.79&36.78\\
0.3&0.35&0.35&0.3&0.5&0.2&28.30&40.67\\
0.3&0.35&0.35&0.3&0.6&0.1&25.16&37.23\\
0.3&0.35&0.35&1&0&0&26.42&38.00\\
0.4&0.3&0.3&0.15&0.55&0.3&25.16&38.50\\
0.4&0.3&0.3&0.2&0.55&0.25&25.79&38.37\\
0.4&0.3&0.3&0.3&0.5&0.2&27.67&41.06\\
0.4&0.3&0.3&0.3&0.6&0.1&25.79&38.58\\
0.4&0.3&0.3&1&0&0&23.27&35.46\\
1&0&0&0.15&0.55&0.3&27.04&39.47\\
1&0&0&0.2&0.55&0.25&28.30&38.12\\
1&0&0&0.3&0.5&0.2&24.53&37.02\\
1&0&0&0.3&0.6&0.1&26.42&35.89\\
1&0&0&1&0&0&24.53&37.76\\
\thickhline
\end{tabular}
\caption{An elaborate overview of HGOT+KNN's various hyperparameter combinations being explored, along with their corresponding EM and F1 scores, within the medium-length category of the Open-SQuAD dataset.}
\label{tab:open-squad_grid_search}
\end{table}

\begin{table}[H]
\centering
\small
\begin{tabular}{p{0.1\linewidth}|p{0.1\linewidth}|p{0.1\linewidth}|p{0.1\linewidth}|p{0.1\linewidth}|p{0.1\linewidth}|p{0.1\linewidth}}
\thickhline
\bf $\alpha$&\bf $\beta$&\bf $\gamma$&\bf $w_1$&\bf $w_2$&\bf $w_3$&\bf EM\\
\hline
0.1&0.45&0.45&0.15&0.55&0.3&53.33\\
0.1&0.45&0.45&0.2&0.55&0.25&54.00\\
0.1&0.45&0.45&0.3&0.5&0.2&57.33\\
0.1&0.45&0.45&0.3&0.6&0.1&54.67\\
0.1&0.45&0.45&1&0&0&61.33\\
0.2&0.4&0.4&0.15&0.55&0.3&51.33\\
0.2&0.4&0.4&0.2&0.55&0.25&56.67\\
0.2&0.4&0.4&0.3&0.5&0.2&52.00\\
0.2&0.4&0.4&0.3&0.6&0.1&59.33\\
0.2&0.4&0.4&1&0&0&57.33\\
0.3&0.35&0.35&0.15&0.55&0.3&57.33\\
0.3&0.35&0.35&0.2&0.55&0.25&57.33\\
0.3&0.35&0.35&0.3&0.5&0.2&61.33\\
0.3&0.35&0.35&0.3&0.6&0.1&56.67\\
0.3&0.35&0.35&1&0&0&61.33\\
0.4&0.3&0.3&0.15&0.55&0.3&59.33\\
0.4&0.3&0.3&0.2&0.55&0.25&56.67\\
0.4&0.3&0.3&0.3&0.5&0.2&60.00\\
0.4&0.3&0.3&0.3&0.6&0.1&56.67\\
0.4&0.3&0.3&1&0&0&60.67\\
1&0&0&0.15&0.55&0.3&58.00\\
1&0&0&0.2&0.55&0.25&58.00\\
1&0&0&0.3&0.5&0.2&54.67\\
1&0&0&0.3&0.6&0.1&52.67\\
1&0&0&1&0&0&58.00\\
\thickhline
\end{tabular}
\caption{A detailed examination of the numerous hyperparameter configurations tested for HGOT+KNN, together with their respective EM scores, specifically within the medium-length category of the FEVER dataset.}
\label{tab:fever_grid_search}
\end{table}

\begin{table}[H]
\centering
\small
\begin{tabular}{p{0.1\linewidth}|p{0.1\linewidth}|p{0.1\linewidth}|p{0.1\linewidth}|p{0.1\linewidth}|p{0.1\linewidth}|p{0.1\linewidth}|p{0.1\linewidth}}
\thickhline
\bf $\alpha$&\bf $\beta$&\bf $\gamma$&\bf $w_1$&\bf $w_2$&\bf $w_3$&\bf EM&\bf F1\\
\hline
0.1&0.45&0.45&0.15&0.55&0.3&42.57&54.49\\
0.1&0.45&0.45&0.2&0.55&0.25&39.19&51.58\\
0.1&0.45&0.45&0.3&0.5&0.2&40.54&52.91\\
0.1&0.45&0.45&0.3&0.6&0.1&39.86&51.94\\
0.1&0.45&0.45&1&0&0&43.92&54.63\\
0.2&0.4&0.4&0.15&0.55&0.3&43.24&55.93\\
0.2&0.4&0.4&0.2&0.55&0.25&39.86&53.81\\
0.2&0.4&0.4&0.3&0.5&0.2&41.22&53.63\\
0.2&0.4&0.4&0.3&0.6&0.1&40.54&52.39\\
0.2&0.4&0.4&1&0&0&43.92&54.63\\
0.3&0.35&0.35&0.15&0.55&0.3&41.89&54.58\\
0.3&0.35&0.35&0.2&0.55&0.25&39.86&53.25\\
0.3&0.35&0.35&0.3&0.5&0.2&41.22&54.17\\
0.3&0.35&0.35&0.3&0.6&0.1&40.54&52.17\\
0.3&0.35&0.35&1&0&0&43.92&54.63\\
0.4&0.3&0.3&0.15&0.55&0.3&41.89&54.58\\
0.4&0.3&0.3&0.2&0.55&0.25&38.51&52.35\\
0.4&0.3&0.3&0.3&0.5&0.2&41.22&53.95\\
0.4&0.3&0.3&0.3&0.6&0.1&40.54&52.79\\
0.4&0.3&0.3&1&0&0&43.92&54.63\\
1&0&0&0.15&0.55&0.3&40.54&54.20\\
1&0&0&0.2&0.55&0.25&39.86&53.47\\
1&0&0&0.3&0.5&0.2&40.54&52.98\\
1&0&0&0.3&0.6&0.1&39.86&53.02\\
1&0&0&1&0&0&43.92&55.08\\
\thickhline
\end{tabular}
\caption{A comprehensive review of the different hyperparameter combinations tested on HGOT+KNN, including both their EM and F1 scores, within the medium-length category of the HotPotQA dataset.}
\label{tab:hotpotqa_grid_search}
\end{table}

\end{document}